\documentclass{article}

\usepackage[final]{neurips_2025}

\usepackage[utf8]{inputenc}
\usepackage{CJKutf8}

\usepackage[T1]{fontenc}
\usepackage{hyperref}
\usepackage{url}
\usepackage{booktabs}
\usepackage{amsfonts}
\usepackage{amsmath}
\usepackage{amssymb}
\usepackage{nicefrac}
\usepackage{subcaption}
\usepackage{microtype}
\usepackage{xcolor}
\usepackage{graphicx}
\usepackage{algorithm}
\usepackage{algorithmic}

\title{Embedding Inversion via Conditional Masked Diffusion Language Models}

\makeatletter
\renewcommand{\@noticestring}{}
\makeatother

\author{%
  Han Xiao \\
  Jina AI by Elastic \\
  \texttt{han.xiao@jina.ai} \\
}

\begin{document}

\maketitle

\begin{abstract}
We frame embedding inversion as conditional masked diffusion, recovering all tokens in parallel through iterative denoising rather than sequential autoregressive generation. A masked diffusion language model is conditioned on the target embedding via adaptive layer normalization, requiring only 8 forward passes with no access to the target encoder at inference time. On 32-token sequences across three embedding models, the method achieves token recovery through parallel denoising without requiring encoder access, iterative correction, or architecture-specific alignment. Source code and live demo are available at \url{https://github.com/jina-ai/embedding-inversion-demo}.
\end{abstract}

\section{Introduction}

Text embeddings power modern retrieval systems, and production deployments routinely treat them as safe, anonymized representations. Vec2Text~\citep{morris2023text} challenged this assumption by recovering 92\% of 32-token sequences from their embeddings using a T5 encoder-decoder with iterative correction. Subsequent work has expanded the attack surface: ALGEN~\citep{algen} enables cross model inversion with few-shot alignment, and Zero2Text~\citep{zero2text} achieves training free inversion via LLM priors and online regression.

These methods share a common design: they generate tokens autoregressively, then iteratively re-embed the hypothesis to compute a correction signal. This creates two practical bottlenecks. First, each correction step requires a forward pass through the target embedding model, making the attack cost proportional to the number of iterations. Vec2Text typically requires over 20 iterations per sequence. Second, the autoregressive backbone accumulates errors left-to-right, with no mechanism to revise earlier tokens based on later context.

We propose an alternative formulation: embedding inversion as \textit{conditional masked diffusion}. Starting from a fully masked sequence, a denoising model iteratively reveals tokens at all positions in parallel, conditioned on the target embedding vector via adaptive layer normalization. Correction is built into the diffusion process itself: each step refines all positions simultaneously using global context, without ever reembedding the current hypothesis. The embedding vector enters only through AdaLN modulation, making the approach encoder agnostic: the same architecture applies to any embedding model without alignment training. We validate on three encoders with different architectures and dimensionalities, with no access to the target encoder at inference time.

\begin{figure}[!htbp]
\centering
\includegraphics[width=\textwidth]{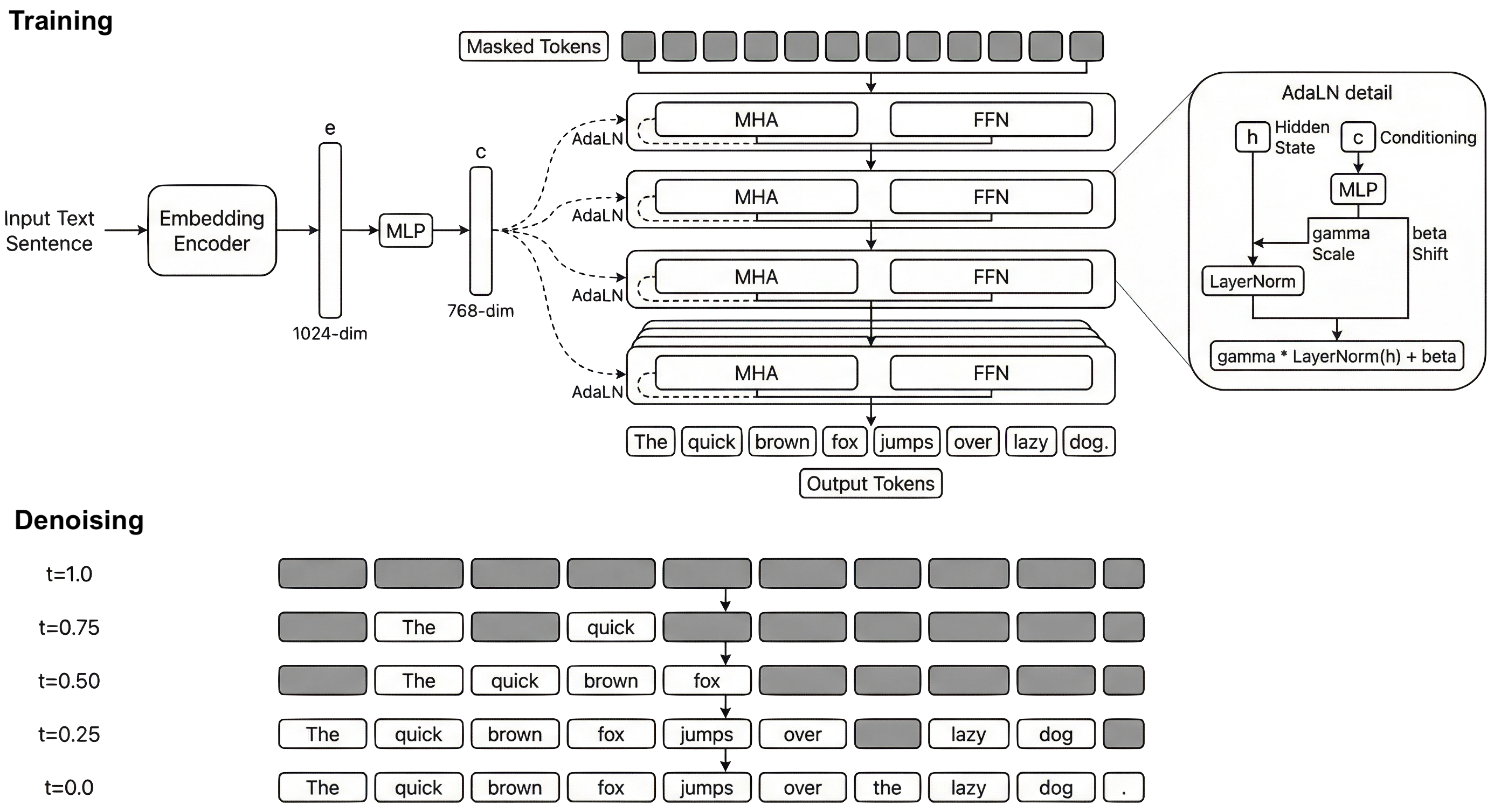}
\caption{Architecture of the Conditional Masked Diffusion Language Model. The embedding vector is projected and injected into each transformer layer via AdaLN conditioning. The model predicts original tokens at masked positions through iterative denoising.}
\label{fig:architecture}
\end{figure}

\section{Related Work}

\subsection{Embedding Inversion Attacks}

Embedding inversion emerged as a research area with Vec2Text~\citep{morris2023text}, which demonstrated that T5 encoder-decoder models could recover 92\% exact matches on 32-token sequences through hypothesis generation followed by iterative correction. The correction mechanism computes embedding distances and refines outputs through multiple forward passes, but requires compatible embedding architectures and suffers from autoregressive error accumulation. 

The field has advanced rapidly with methods addressing Vec2Text's architectural constraints. ALGEN~\citep{algen} introduced few-shot cross model alignment, demonstrating that embedding spaces can be aligned with only 1k training samples through one-step optimization, enabling inversion across incompatible architectures. Zero2Text~\citep{zero2text} achieved training free inversion using LLM priors combined with online ridge regression, eliminating the need for paired training data entirely. On MS MARCO, Zero2Text reported 1.8$\times$ ROUGE-L improvement over baselines in black-box cross-domain settings. These methods show that embedding inversion generalizes across architectures and data regimes. Our work contributes a diffusion-based approach, replacing sequential generation and explicit correction with parallel masked denoising.

\subsection{Discrete Diffusion Models}

Discrete diffusion began with D3PM~\citep{austin2021structured}, which extended continuous diffusion to categorical distributions through absorbing state processes. Masked Diffusion Language Models~\citep{sahoo2024simple} simplified this framework by using uniform masking with log-linear noise schedules, achieving competitive language modeling performance while enabling parallel generation. The field has since diversified: Score Entropy Discrete Diffusion~\citep{sedd} introduced entropy-based scoring, providing improved sample quality through better noise scheduling. Constrained Discrete Diffusion~\citep{cdd} added constraint satisfaction mechanisms for controlled generation tasks.

Our conditional MDLM builds on this foundation, adapting masked diffusion to the embedding inversion task through adaptive layer normalization conditioning.

\subsection{Conditional Diffusion}

Conditioning mechanisms for diffusion models have evolved primarily in continuous domains. Classifier-free guidance~\citep{ho2022classifierfree} enables conditional generation by training a single model with dropped conditioning signals, then interpolating predictions at inference. Classifier guidance~\citep{dhariwal2021diffusion} uses external classifier gradients to steer generation toward desired attributes. For vision tasks, Diffusion Transformers~\citep{peebles2023scalable} introduced adaptive layer normalization that modulates layer normalization parameters based on conditioning signals, providing fine-grained control over feature representations at each transformer layer. We adapt AdaLN to discrete text generation, using it to inject embedding information into each denoising step. This conditioning mechanism is architecture agnostic, working with any embedding model without requiring alignment training or model specific modifications, in contrast to Vec2Text's T5-specific architecture or ALGEN's explicit alignment procedure.

\section{Method}

We use the following notation throughout: $\mathbf{x} = (x_1, \ldots, x_n)$ denotes a token sequence of length $n$ from vocabulary $\mathcal{V}$; $\mathbf{e} \in \mathbb{R}^d$ denotes the embedding vector; $t \in [0,1]$ denotes the diffusion timestep with $t=0$ being fully unmasked and $t=1$ being fully masked; $\theta$ denotes the model parameters; $\mathbf{c} \in \mathbb{R}^{D_h}$ denotes the projected conditioning vector with hidden dimension $D_h$; $x_t$ denotes the masked sequence at timestep $t$; $x_0$ denotes the original unmasked sequence.

\subsection{Problem Formulation}

Given an embedding function $f: \mathcal{V}^n \to \mathbb{R}^d$ and embedding vector $\mathbf{e} = f(\mathbf{x})$, we seek to recover the original sequence by maximizing the conditional probability:
\begin{equation}
\hat{\mathbf{x}} = \arg\max_{\mathbf{x}'} p_\theta(\mathbf{x}' | \mathbf{e})
\end{equation}
where $p_\theta(\mathbf{x} | \mathbf{e})$ is modeled using masked diffusion with adaptive layer normalization conditioning.

\subsection{Masked Diffusion Process}

Following MDLM~\citep{sahoo2024simple}, we define a forward noising process that gradually masks tokens according to a noise schedule. For each token position $i$ at timestep $t$, the forward transition is:
\begin{equation}
q(x_{t,i} | x_{0,i}) = \begin{cases}
x_{0,i} & \text{with probability } \alpha_t \\
[\text{MASK}] & \text{with probability } 1 - \alpha_t
\end{cases}
\label{eq:forward}
\end{equation}
where $x_{t,i}$ is the token at position $i$ and timestep $t$, $x_{0,i}$ is the original token, and $\alpha_t$ is the survival probability. We use the log-linear schedule $\alpha_t = e^{-\lambda t}$ with $\lambda = 5.0$, which concentrates masking in later timesteps while preserving structure in early denoising stages. The reverse process learns to predict the original token $x_{0,i}$ at each masked position given the partially masked sequence $x_t$, timestep $t$, and conditioning embedding $\mathbf{e}$. The model outputs a categorical distribution over the vocabulary:
\begin{equation}
p_\theta(x_{0,i} | x_t, t, \mathbf{e}) = \text{Categorical}(\text{softmax}(\mathbf{z}_i))
\label{eq:reverse}
\end{equation}
where $\mathbf{z}_i \in \mathbb{R}^{|\mathcal{V}|}$ are the logits for position $i$ produced by the transformer network parameterized by $\theta$. The model predicts all positions in parallel, conditioned on the global context provided by the embedding. We minimize the Rao-Blackwellized ELBO with $1/t$ weighting:
\begin{equation}
\mathcal{L}(\theta) = \mathbb{E}_{t \sim \text{Uniform}[0,1]} \mathbb{E}_{\mathbf{x}_0 \sim \mathcal{D}} \mathbb{E}_{x_t \sim q(x_t|x_0)} \left[ \frac{1}{t} \sum_{i: x_{t,i} = [\text{MASK}]} -\log p_\theta(x_{0,i} | x_t, t, \mathbf{e}) \right]
\label{eq:loss}
\end{equation}
where $\mathcal{D}$ is the data distribution, the sum is over masked positions only, and the $1/t$ weighting upweights the low-noise regime ($t \to 0$), where few tokens remain masked and precise reconstruction matters most.

\subsection{Model Architecture}

Our model consists of three components: embedding projection, transformer backbone, and adaptive layer normalization conditioning (Figure~\ref{fig:architecture}). The input embedding $\mathbf{e} \in \mathbb{R}^d$, where $d$ is determined by the target encoder, is projected to the transformer hidden dimension $D_h$ via a two-layer MLP:
\begin{equation}
\mathbf{c} = \mathbf{W}_2 \cdot \text{GELU}(\mathbf{W}_1 \mathbf{e} + \mathbf{b}_1) + \mathbf{b}_2
\label{eq:projection}
\end{equation}
where $\mathbf{W}_1 \in \mathbb{R}^{D_h \times d}$, $\mathbf{W}_2 \in \mathbb{R}^{D_h \times D_h}$, and $\mathbf{b}_1, \mathbf{b}_2 \in \mathbb{R}^{D_h}$ are learned parameters. We primarily use a 22-layer transformer initialized from multilingual BERT with vocabulary size 256K. With $D_h = 768$ and FFN dimension 3072, the model has 388M total parameters. When freezing the pretrained backbone, 191M parameters remain trainable (49.3\%), consisting of the embedding projection MLP, AdaLN conditioning MLPs, and output layer. We also experiment with 8-layer (268M parameters) and 2-layer (217M parameters with 20M trainable when frozen) configurations to assess depth-accuracy tradeoffs. Input and output embeddings are weight-tied to reduce parameters given the large vocabulary size.

Following DiT~\citep{peebles2023scalable}, we condition each transformer layer on both the timestep $t$ and the embedding vector $\mathbf{c}$ via adaptive layer normalization. For each layer $\ell$, we compute modulation parameters:
\begin{align}
\gamma_t^{(\ell)}, \beta_t^{(\ell)} &= \text{MLP}_t^{(\ell)}(t) \label{eq:adaln-time} \\
\gamma_c^{(\ell)}, \beta_c^{(\ell)} &= \text{MLP}_c^{(\ell)}(\mathbf{c}) \label{eq:adaln-cond} \\
\gamma^{(\ell)} &= \gamma_t^{(\ell)} + \gamma_c^{(\ell)} \label{eq:adaln-scale} \\
\beta^{(\ell)} &= \beta_t^{(\ell)} + \beta_c^{(\ell)} \label{eq:adaln-shift}
\end{align}
where $\text{MLP}_t^{(\ell)}$ and $\text{MLP}_c^{(\ell)}$ are single-layer MLPs that output vectors of dimension $D_h$, $\gamma^{(\ell)}$ is the scale parameter, and $\beta^{(\ell)}$ is the shift parameter. The layer normalization at layer $\ell$ is then modulated:
\begin{equation}
\text{AdaLN}(\mathbf{h}^{(\ell)}) = \gamma^{(\ell)} \odot \frac{\mathbf{h}^{(\ell)} - \mu(\mathbf{h}^{(\ell)})}{\sigma(\mathbf{h}^{(\ell)})} + \beta^{(\ell)}
\label{eq:adaln}
\end{equation}
where $\mathbf{h}^{(\ell)} \in \mathbb{R}^{n \times D_h}$ is the input to layer $\ell$, $\mu(\cdot)$ and $\sigma(\cdot)$ compute mean and standard deviation over the hidden dimension, and $\odot$ denotes element-wise multiplication. This formulation allows the conditioning signal and timestep to independently modulate the layer normalization at each depth, providing fine-grained control over feature representations.

\subsection{Decoding Strategies}

We consider five strategies for generating tokens from the trained model.

\textbf{Sequential greedy decoding} unmasks tokens left to right:
\begin{equation}
x_i = \arg\max_{v \in \mathcal{V}} p_\theta(v \mid x_{<i}, [\text{MASK}]^{n-i}, \mathbf{e}, t)
\label{eq:sequential}
\end{equation}
where $t = (n-i)/n$ is the fraction of remaining masked tokens. This produces coherent text but sacrifices the parallel nature of diffusion.

\textbf{Euler sampling} applies the Euler method to the reverse diffusion process, starting from $x_1 = [\text{MASK}]^n$ with uniform timesteps from $t{=}1$ to $t{=}0$:
\begin{equation}
\hat{x}_{0,i} \sim p_\theta(x_{0,i} \mid x_t, t, \mathbf{e}) \quad \forall\, i
\label{eq:euler}
\end{equation}
sampling all positions simultaneously at each step.

\textbf{Euler with remasking} adds a correction mechanism: after each Euler step, a fraction $\tau$ of positions with the lowest confidence $\max_v p_\theta(v \mid x_t, t, \mathbf{e})$ are re-masked:
\begin{equation}
x_{t',i} = \begin{cases} \hat{x}_{0,i} & \text{if position } i \text{ is not in the bottom-}\tau \text{ fraction} \\ [\text{MASK}] & \text{otherwise} \end{cases}
\label{eq:remasking}
\end{equation}
allowing subsequent steps to refine uncertain predictions. We find $\tau = 0.05$ optimal (Table~\ref{tab:remask}).

\textbf{Confidence-based decoding} adapts the MaskGIT approach to text generation. At each iteration step $s$, we predict all token positions and unmask the top-$k_s$ most confident tokens:
\begin{align}
c_i &= \max_{v \in \mathcal{V}} p_\theta(v \mid x_t, t, \mathbf{e}) \quad \forall\, i \label{eq:confidence} \\
x_{t',i} &= \begin{cases} \arg\max_{v \in \mathcal{V}} p_\theta(v \mid x_t, t, \mathbf{e}) & \text{if } c_i \text{ is in top-}k_s \text{ confidence scores} \\ [\text{MASK}] & \text{otherwise} \end{cases}
\label{eq:confidence-decode}
\end{align}
where $c_i$ is the confidence at position $i$, and $k_s$ decreases from $n$ to 0 over a fixed number of iterations. This allows the model to first commit to high-confidence predictions and progressively fill uncertain positions using the context from already decoded tokens.

\textbf{Two-stage decoding} first generates a hypothesis via sequential greedy decoding, then refines it using Euler sampling initialized at this hypothesis rather than a fully masked sequence.

\section{Experimental Results}

We train on 2M samples from C4~\citep{raffel2020exploring}, filtered to 32 tokens. Training uses batch size 380-400 for up to 200K steps with AdamW optimizer at learning rate $10^{-4}$, warmup of 2000 steps, and EMA decay 0.9999. We employ a log-linear noise schedule with $\lambda = 5.0$ following~\citet{sahoo2024simple}. Timesteps are sampled uniformly from $[0, 1]$. Embeddings are computed using the target encoder and cached. We evaluate on three embedding models with different architectures and dimensionalities: jina-embeddings-v3~\citep{jina2024embeddings} with 570M parameters and 1024-dimensional embeddings, Qwen3-Embedding-0.6B with 600M parameters and 1024-dimensional embeddings, and EmbeddingGemma-300m with 300M parameters and 768-dimensional embeddings. We train separate models for each encoder using multilingual data from mC4 to assess generalization across embedding spaces.

The 22-layer configuration with 256K vocabulary has 388M total parameters. When freezing the pretrained mmBERT backbone, 191M parameters (49.3\%) remain trainable, consisting of the embedding projection MLP, AdaLN conditioning MLPs, and output layer. The 8-layer configuration has 268M total parameters, all trainable. The 2-layer configuration has 217M total parameters with 20M trainable when frozen.

Table~\ref{tab:encoder-comparison} shows results across all three embedding encoders using sequential greedy decoding. These numbers represent training accuracy on the cached embedding dataset and may not reflect generalization to held-out test data. The reported accuracies vary by encoder: Qwen3-Embedding at 81.3\%, EmbeddingGemma at 78.8\%, and jina-v3 at 76.0\%. All models are trained on multilingual data from mC4. Current ablation experiments with the 22-layer mmBERT backbone show training accuracies of 68.2\% for the no-freeze configuration and 63.8\% for the freeze configuration, indicating substantial variance across experimental setups and checkpoints.

\begin{table}[!htbp]
\centering
\caption{Performance across embedding encoders using sequential greedy decoding. All trained on 2M multilingual samples from mC4. Accuracies are training metrics on cached embeddings and should not be interpreted as test-set performance. Best checkpoint selected by validation loss.}
\label{tab:encoder-comparison}
\begin{tabular}{lccccc}
\toprule
Encoder & Token Acc. & Steps & Val Loss & Vocab & Embed Dim \\
\midrule
Qwen3-Embedding-0.6B & 81.3\% & 72.5K & 1.317 & 152K & 1024 \\
EmbeddingGemma-300m & 78.8\% & 49.5K & 1.55 & 262K & 768 \\
jina-embeddings-v3 & 76.0\% & 62.5K & 1.60 & 250K & 1024 \\
\bottomrule
\end{tabular}
\end{table}

Table~\ref{tab:main-results} compares decoding strategies across all three encoders on 10 languages. Cosine similarity is averaged over the same sentence translated into English, Chinese, German, Japanese, French, Spanish, Korean, Russian, Arabic, and Portuguese. Sequential greedy decoding shows the highest similarity for jina-v3 and EmbeddingGemma, while two-stage decoding performs best for Qwen3-Embedding.

\begin{table}[!htbp]
\centering
\caption{Average cosine similarity across decoding strategies and encoders, evaluated on 10 languages per encoder.}
\label{tab:main-results}
\begin{tabular}{lccc}
\toprule
Decoding Method & jina-embeddings-v3 & Qwen3-Embedding & EmbeddingGemma \\
\midrule
Sequential Greedy & \textbf{0.715} & 0.585 & \textbf{0.621} \\
Euler Sampling & 0.667 & 0.556 & 0.604 \\
Euler + Remasking & 0.665 & 0.584 & 0.595 \\
Two-Stage & 0.667 & \textbf{0.591} & 0.605 \\
\bottomrule
\end{tabular}
\end{table}

Euler with remasking at 0.05 improves over vanilla Euler by 2.6 percentage points in token accuracy. Two-stage decoding achieves the highest exact match rate at 13.1\%. Baselines confirm that embedding conditioning is necessary: random tokens achieve 0.02\% accuracy, while an unconditional language model achieves 2.1\% accuracy despite high fluency with BLEU score 89.3, indicating that fluency alone does not ensure faithful reconstruction.

Table~\ref{tab:remask} shows optimal performance at remask probability 0.05 for Euler sampling with adaptive remasking. Higher rates discard correct predictions, lower rates provide insufficient correction.

\begin{table}[!htbp]
\centering
\caption{Effect of remasking probability on Euler sampling performance.}
\label{tab:remask}
\begin{tabular}{lccc}
\toprule
Re-mask Prob. & Token Acc. & Cosine Sim. & BLEU \\
\midrule
0.00 (no re-mask) & 65.2\% & 0.81 & 38.7 \\
0.05 & \textbf{67.8\%} & \textbf{0.82} & \textbf{42.1} \\
0.10 & 66.3\% & 0.81 & 40.2 \\
0.20 & 63.7\% & 0.80 & 37.1 \\
\bottomrule
\end{tabular}
\end{table}

\begin{figure}[!htbp]
\centering
\begin{subfigure}[t]{0.49\textwidth}
\includegraphics[width=\textwidth]{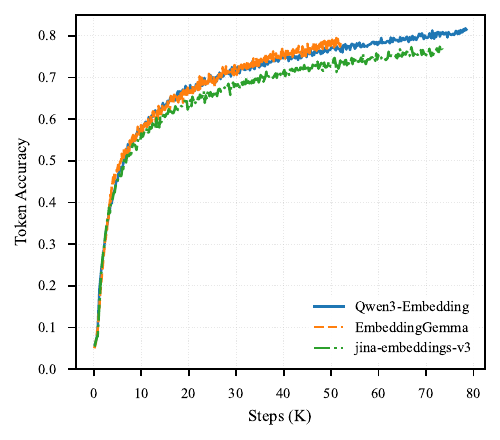}
\caption{Token accuracy}
\end{subfigure}
\hfill
\begin{subfigure}[t]{0.49\textwidth}
\includegraphics[width=\textwidth]{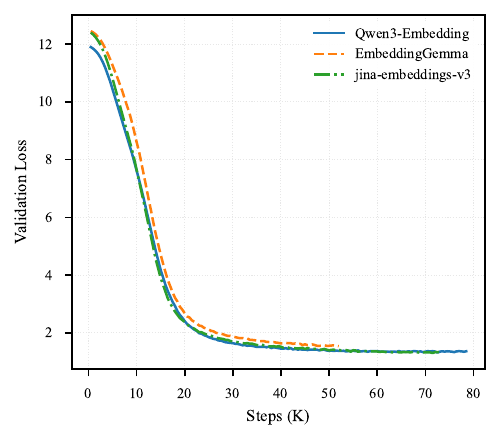}
\caption{Validation loss}
\end{subfigure}
\caption{Training dynamics across three embedding encoders on 2M multilingual samples. Qwen3-Embedding reaches 81.3\% training token accuracy at 72.5K steps with validation loss 1.32. All models show plateauing behavior beyond 50K steps, indicating that further improvements may require architectural changes rather than extended training.}
\label{fig:training-curves}
\end{figure}

\section{Conclusion and Future Work}

We presented embedding inversion via conditional masked diffusion, demonstrating that parallel denoising can recover tokens from embedding vectors without access to the target encoder. The method operates in a strictly black-box setting: the embedding vector is the only input at inference time, with no correction loop through the encoder. This threat model is more restrictive than Vec2Text, which requires encoder access for iterative refinement, or ALGEN, which needs alignment training on encoder outputs.

This architectural difference also explains the performance gap. Autoregressive methods with iterative correction benefit from a closed-loop feedback signal. Each refinement step re-embeds the current hypothesis and computes an explicit error signal against the target. Diffusion-based inversion lacks this mechanism. The embedding conditions the generation through AdaLN modulation but never verifies whether the output actually maps back to the target embedding. Errors compound without correction, which our ablation results confirm: dynamic masking schedules that provide implicit curriculum learning substantially outperform fixed schedules, partially compensating for the absence of explicit feedback.

Two directions appear promising for closing this gap. Classifier-free guidance could inject stronger embedding signal during sampling without requiring encoder access at inference time. Alternatively, combining diffusion-based initialization with a lightweight correction step through a distilled or approximate encoder could recover much of the closed-loop advantage while maintaining broader applicability across encoder architectures. Scaling to longer sequences via hierarchical diffusion over sentence-level chunks is a natural next step.

\bibliographystyle{plainnat}
\bibliography{references}

\begin{thebibliography}{12}
\providecommand{\natexlab}[1]{#1}
\providecommand{\url}[1]{\texttt{#1}}
\expandafter\ifx\csname urlstyle\endcsname\relax
  \providecommand{\doi}[1]{doi: #1}\else
  \providecommand{\doi}{doi: \begingroup \urlstyle{rm}\Url}\fi

\bibitem[Austin et~al.(2021)Austin, Johnson, Ho, Tarlow, and van~den
  Berg]{austin2021structured}
Jacob Austin, Daniel~D. Johnson, Jonathan Ho, Daniel Tarlow, and Rianne van~den
  Berg.
\newblock Structured denoising diffusion models in discrete state-spaces.
\newblock In \emph{NeurIPS}, volume~34, pages 17981--17993, 2021.

\bibitem[Cardei et~al.(2025)Cardei, Christopher, Hartvigsen, Bartoldson,
  Kailkhura, and Fioretto]{cdd}
Michael Cardei, Jacob~K. Christopher, Thomas Hartvigsen, Brian~R. Bartoldson,
  Bhavya Kailkhura, and Ferdinando Fioretto.
\newblock Constrained language generation with discrete diffusion models.
\newblock \emph{arXiv preprint arXiv:2503.09790}, 2025.

\bibitem[Chen et~al.(2025)Chen, Xu, and Bjerva]{algen}
Yiyi Chen, Qiongkai Xu, and Johannes Bjerva.
\newblock Algen: Few-shot inversion attacks on textual embeddings via
  cross-model alignment and generation.
\newblock In \emph{ACL}, 2025.

\bibitem[Dhariwal and Nichol(2021)]{dhariwal2021diffusion}
Prafulla Dhariwal and Alexander Nichol.
\newblock Diffusion models beat gans on image synthesis.
\newblock In \emph{NeurIPS}, volume~34, pages 8780--8794, 2021.

\bibitem[Ho and Salimans(2022)]{ho2022classifierfree}
Jonathan Ho and Tim Salimans.
\newblock Classifier-free diffusion guidance.
\newblock In \emph{NeurIPS 2021 Workshop on Deep Generative Models and
  Downstream Applications}, 2022.

\bibitem[Kim et~al.(2026)Kim, Kang, Lee, Baek, and Kang]{zero2text}
Doohyun Kim, Donghwa Kang, Kyungjae Lee, Hyeongboo Baek, and Brent~Byunghoon
  Kang.
\newblock Zero2text: Zero-training cross-domain inversion attacks on textual
  embeddings.
\newblock \emph{arXiv preprint arXiv:2602.01757}, 2026.

\bibitem[Lou et~al.(2024)Lou, Meng, and Ermon]{sedd}
Aaron Lou, Chenlin Meng, and Stefano Ermon.
\newblock Discrete diffusion modeling by estimating the ratios of the data
  distribution.
\newblock In \emph{ICML}, pages 32819--32848, 2024.

\bibitem[Morris et~al.(2023)Morris, Kuleshov, Shmatikov, and
  Rush]{morris2023text}
John~X. Morris, Volodymyr Kuleshov, Vitaly Shmatikov, and Alexander~M. Rush.
\newblock Text embeddings reveal (almost) as much as text.
\newblock In \emph{EMNLP}, pages 12448--12460, 2023.

\bibitem[Peebles and Xie(2023)]{peebles2023scalable}
William Peebles and Saining Xie.
\newblock Scalable diffusion models with transformers.
\newblock In \emph{ICCV}, pages 4195--4205, 2023.

\bibitem[Raffel et~al.(2020)Raffel, Shazeer, Roberts, Lee, Narang, Matena,
  Zhou, Li, and Liu]{raffel2020exploring}
Colin Raffel, Noam Shazeer, Adam Roberts, Katherine Lee, Sharan Narang, Michael
  Matena, Yanqi Zhou, Wei Li, and Peter~J. Liu.
\newblock Exploring the limits of transfer learning with a unified text-to-text
  transformer.
\newblock \emph{Journal of Machine Learning Research}, 21\penalty0
  (140):\penalty0 1--67, 2020.

\bibitem[Sahoo et~al.(2024)Sahoo, Arriola, Schiff, Gokaslan, Marroquin, Chiu,
  Rush, and Kuleshov]{sahoo2024simple}
Subham~S. Sahoo, Marianne Arriola, Yair Schiff, Aaron Gokaslan, Edgar
  Marroquin, Justin~T. Chiu, Alexander Rush, and Volodymyr Kuleshov.
\newblock Simple and effective masked diffusion language models.
\newblock In \emph{NeurIPS}, volume~37, 2024.

\bibitem[Sturua et~al.(2025)Sturua, Mohr, Akram, G{\"u}nther, Wang, Krimmel,
  Wang, Mastrapas, Koukounas, Wang, and Xiao]{jina2024embeddings}
Saba Sturua, Isabelle Mohr, Mohammad~Kalim Akram, Michael G{\"u}nther, Bo~Wang,
  Markus Krimmel, Feng Wang, Georgios Mastrapas, Andreas Koukounas, Nan Wang,
  and Han Xiao.
\newblock jina-embeddings-v3: Multilingual embeddings with task lora.
\newblock In \emph{ECIR}, 2025.

\end{thebibliography}

\appendix

\section{Mask Ratio Ablation with mmBERT Decoder}
\label{app:mask-ablation}

We conduct a separate ablation study using a different decoder architecture to investigate the effect of fixed mask ratios on training dynamics. Unlike the main experiments, which use an 8-layer transformer trained from scratch with the GPT-2 tokenizer, these experiments use a 22-layer transformer initialized from mmBERT pretrained weights with a 256K multilingual vocabulary. Five models were trained with fixed mask percentages of 10\%, 20\%, 40\%, 80\%, and 100\%, sharing identical architecture (388M parameters), training configuration (batch size 380, learning rate $10^{-4}$, warmup 2000 steps), and pretrained initialization.

\begin{table}[!htbp]
\centering
\caption{Mask ratio ablation with mmBERT decoder. All models early-stopped before 200K steps.}
\label{tab:mask-ablation}
\begin{tabular}{lcccc}
\toprule
Mask \% & Best Step & Best Val Loss & Final Train Acc & Tokens (M) \\
\midrule
10\%  & 43,500 & 5.9943 & 30.2\% & 3,413 \\
20\%  & 42,500 & 5.9197 & 21.0\% & 3,284 \\
40\%  & 38,500 & 5.8730 & 14.0\% & 3,174 \\
80\%  & 35,500 & 5.7933 & 10.8\% & 2,955 \\
100\% & 30,500 & 5.8410 & 10.1\% & 2,590 \\
\bottomrule
\end{tabular}
\end{table}

Higher mask ratios achieve lower validation loss, with 80\% masking reaching 5.79 (Table~\ref{tab:mask-ablation}). Training accuracy inversely correlates with mask ratio as expected. Higher mask ratios also converge faster, with 100\% masking reaching its best validation loss at step 30,500 versus 43,500 for 10\% masking. However, all fixed-mask models underperform configurations using the log-linear schedule from the main experiments, suggesting that dynamic masking provides curriculum learning benefits.

\begin{figure}[!htbp]
\centering
\begin{subfigure}[t]{0.49\textwidth}
\includegraphics[width=\textwidth]{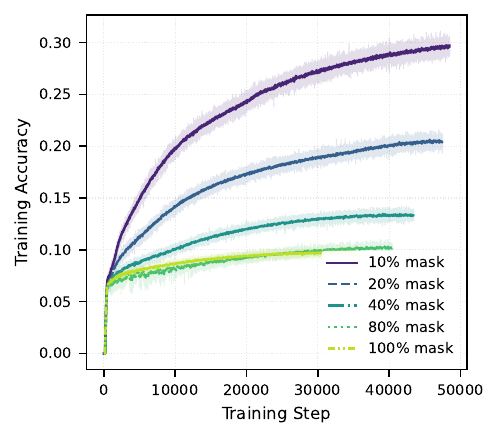}
\caption{Training accuracy curves}
\end{subfigure}
\hfill
\begin{subfigure}[t]{0.49\textwidth}
\includegraphics[width=\textwidth]{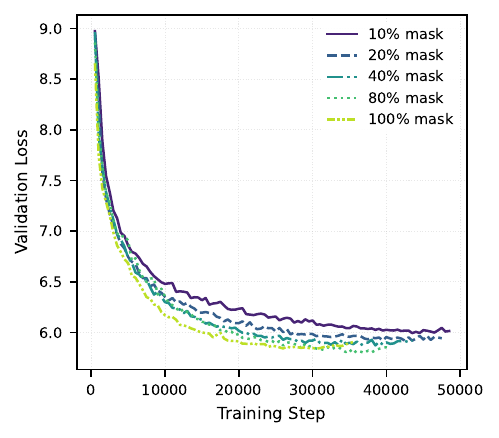}
\caption{Validation loss curves}
\end{subfigure}
\caption{Training dynamics across fixed mask ratios with mmBERT decoder. All models plateau early, and higher training accuracy from lower mask ratios does not translate into better validation performance.}
\label{fig:mask-ablation}
\end{figure}

\end{document}